\documentclass[a4paper, 10pt, conference]{ieeeconf}      
\usepackage{FG2024}
\FGfinalcopy 

\IEEEoverridecommandlockouts                      
\usepackage{times}
\usepackage{epsfig}
\usepackage{graphicx}
\usepackage{amsmath}
\usepackage{amssymb}
\usepackage{amssymb}
\usepackage{amsfonts}
\usepackage{multirow}
\usepackage{tabularx}
\usepackage{mathrsfs}
\usepackage{booktabs}
\usepackage{multirow}
\usepackage{graphicx}
\usepackage{float}
\usepackage{siunitx}

\usepackage[capitalize]{cleveref}
\crefname{section}{Sec.}{Secs.}
\Crefname{section}{Section}{Sections}
\Crefname{table}{Table}{Tables}
\crefname{table}{Tab.}{Tabs.}

\def\FGPaperID{232} 

\title{Face to Cartoon Incremental Super-Resolution using Knowledge Distillation}

\author{Trinetra Devkatte$^{\ast}$, Shiv Ram Dubey$^{\ast}$, Satish Kumar Singh$^{\ast}$, Abdenour Hadid$^{\dagger}$\\
$^{\ast}$Computer Vision and Biometrics Lab, Indian Institute of Information Technology, Allahabad\\
$^{\dagger}$Sorbonne Centre for Artificial Intelligence, Sorbonne University Abu Dhabi\\
{\tt\small mit2021096@iiita.ac.in, srdubey@iiita.ac.in, sk.singh@iiita.ac.in, abdenour.hadid@ieee.org}
}

\begin{document}

\ifFGfinal
\thispagestyle{empty}
\pagestyle{empty}
\else
\author{Anonymous FG2024 submission\\ Paper ID \FGPaperID \\}
\pagestyle{plain}
\fi
\maketitle

\begin{abstract}
Facial super-resolution/hallucination is an important area of research that seeks to enhance low-resolution facial images for a variety of applications. While Generative Adversarial Networks (GANs) have shown promise in this area, their ability to adapt to new, unseen data remains a challenge. This paper addresses this problem by proposing an incremental super-resolution using GANs with knowledge distillation (ISR-KD) for face to cartoon. Previous research in this area has not investigated incremental learning, which is critical for real-world applications where new data is continually being generated. The proposed ISR-KD aims to develop a novel unified framework for facial super-resolution that can handle different settings, including different types of faces such as cartoon face and various levels of detail. To achieve this, a GAN-based super-resolution network was pre-trained on the CelebA dataset and then incrementally trained on the iCartoonFace dataset, using knowledge distillation to retain performance on the CelebA test set while improving the performance on iCartoonFace test set. Our experiments demonstrate the effectiveness of knowledge distillation in incrementally adding capability to the model for cartoon face super-resolution while retaining the learned knowledge for facial hallucination tasks in GANs. 
\end{abstract}

\section{Introduction}
Facial super-resolution/hallucination is a crucial field of research that aims to enhance the quality of low-resolution facial images for various applications, including security systems, medical imaging, and entertainment \cite{wang2020deep}. Generative Adversarial Networks (GANs) \cite{goodfellow2020generative} have shown promising results in facial super-resolution tasks \cite{zhang2020supervised}. However, one of the challenges faced by GANs is with their limited ability to adapt to new and unseen data. This limitation becomes particularly critical in real-world scenarios where new facial data is continuously generated, such as in surveillance systems or video streaming platforms.

In this work, we address this challenge by exploring the potential of incremental learning in the context of GAN-based facial super-resolution. Incremental learning enables the model to continually learn and adapt to new data while retaining knowledge acquired from previous training stages~\cite{wu2019large}. 
The proposed Incremental Super-Resolution with Knowledge Distillation (ISR-KD) aims to develop a unified framework for facial hallucination that can handle various settings, including different types of faces and different levels of detail. The proposed ISR-KD leverages the benefits of knowledge distillation \cite{hinton2015distilling} to retain the performance of the pre-trained GAN-based super-resolution network while incrementally learning from new data.

To evaluate the effectiveness of our approach, we consider a pre-trained GAN-based super-resolution network on the CelebA dataset, which consists of a large collection of celebrity faces, and then incrementally train the network on the iCartoonFace dataset for super-resolution, which contains cartoon-style images. The knowledge distillation loss is utilized for training.
The experimental results demonstrate the effectiveness of knowledge distillation in incrementally expanding the model's capability for facial hallucination tasks within the GAN framework. 
The ability to incrementally adapt to new data and retain previously learned knowledge makes the proposed approach highly suitable for real-world applications where the facial data distribution evolves over time.
The major contributions of this paper are as follows:
\begin{itemize}
    \item This paper proposes an Incremental Super-Resolution technique using Knowledge Distillation (ISR-KD) for exploiting the existing knowledge for super-resolution on new types of images without training from scratch.
    \item The proposed model is originally developed and trained for face super-resolution and incrementally trained for cartoon face super-resolution.
    \item The proposed model is able to improve the cartoon face super-resolution performance with negligible performance drop for original face super-resolution.
\end{itemize}

In the remainder of this paper, Section 2 provides a comprehensive review of literature while Section 3 presents the proposed ISR-KD framework. Sections 4 and 5 describe the experimental setup and the experimental results, respectively. Finally, concluding remarks are drawn in Section 6.

\section{Related Work}
\subsection{Incremental Learning and Knowledge Distillation}
Incremental learning has been widely exploited in computer vision for different applications to incrementally add new classes to a trained model \cite{zhang2020class}. Knowledge distillation is also heavily utilized to transfer the gained knowledge from one model to other model or one type of data to other type of data \cite{gou2021knowledge}, \cite{reddy2023context}.
Welling {\it et al.} \cite{welling2009herding} proposed herding selection criterion to choose samples from the previous dataset. The incremental Classifier and Representation Learning (iCaRL) was proposed by Rebuffi {\it et al.} \cite{8100070} which first extracts features using new data and then performs classification using nearest mean of exemplars rule, after that it combines the classification and distillation loss to adjust the exemplar. End-to-End Incremental Learning (EEIL) was proposed by Castro {\it et al.} \cite{castro2018end} to perform feature extraction and classification. EEIL uses a joint loss function for classification and distillation. The problem of class imbalance was addressed by Hou {\it et al.} \cite{hou2019learning} by proposing a unified classifier and incremental learning. Bias Correction Layer (BiC) was introduced by Wu {\it et al.} \cite{8954008} to handle the last fully connected layer being biased towards new classes.

Chenshen {\it et al.} \cite{wu2018memory} proposed memory replay GANs  which combats catastrophic forgetting problems by joint retraining and aligning replays.  Mengyao {\it et al.} proposed Lifelong GAN \cite{zhai2019lifelong} which used knowledge distillation to combat catastrophic forgetting by encouraging the model to produce visually similar results to a pre-trained model.
The Learning without Forgetting (LwF) \cite{li2017learning} method proposed by Li and Hoiem in 2017 prevents catastrophic forgetting by adding task-specific parameters to the original model when learning a new task. 
However, LwF requires storing parameters for each learned task and is dependent on the correlation between the tasks. Dhar {\it et al.} (2019) proposed Learning without Memorizing (LwM) \cite{dhar2019learning} that allows a model to learn new classes incrementally without requiring data from base classes. This is achieved by restricting the divergence between student and teacher models using attention maps generated from the gradient flow information.

Though incremental learning is widely used to add new classes and knowledge distillation is heavily utilized for light-weight models, they are not well explored for super-resolution over new type of images. In this paper, the incremental learning facilitated by knowledge distillation is exploited for face to cartoon incremental super-resolution.

\subsection{Face Super-Resolution}
In recent years, deep learning models have shown outstanding performance for Face Super-Resolution (FSR) \cite{jiang2021deep}.
Hao {\it et al.} \cite{10.1145/3394171.3413590} proposed PCA-SRGAN which pays attention to the cumulative discrimination in the orthogonal projection space spanned by a PCA projection matrix of face data to improve the performance of GAN-based models on super-resolving face images. 
The Edge and Identity Preserving Network (EIP-Net) \cite{kim2021edge} addresses the distortion of facial components by providing edge information and data distributions. 
A generative and controllable face super-resolution (GCFSR) framework is introduced in \cite{He_2022_CVPR} that reconstructs high-resolution images while preserving identity information without additional priors. A deep FSR method with iterative collaboration between two recurrent networks is proposed in \cite{Ma_2020_CVPR} by leveraging the facial landmarks for image recovery and accurate landmark estimation. A supervised pixel-wise GAN (SPGAN) is investigated in \cite{9132630} that performs the super-resolution at different scales while considering face identity. The denoising diffusion probabilistic models are combined with image-to-image translation to perform super-resolution via repeated refinement \cite{9887996}.    

A pre-prior guided approach is exploited in \cite{9875217} that extracts facial prior information from high-resolution images and embeds them into low-resolution images to improve face reconstruction performance. 
Shuang {\it et al.} \cite{9413117} uses a multi-scale deep network that incorporates both global and local facial priors to generate high-quality super-resolved face images. 
First, the feature extraction module extracts multi-scale features of the input image, then the super-resolution module utilizes these features along with the facial parsing prior to generate high-quality super-resolved face images. 
The 3D facial priors are incorporated into face super-resolution in \cite{10.1007/978-3-030-58548-8_44} by exploiting the facial structures and identity information for improved performance. 

A SPARNet architecture is proposed in \cite{9293182} for face super-resolution by leveraging spatial attention mechanisms to capture key face structures effectively. SPARNet achieves promising performance, even for very low-resolution faces. A self-attention learning network (SLNet) is proposed in \cite{ZENG2023164} for three-stage face hallucination. SLNet leverages the interdependence of low and high-level spaces to achieve better reconstruction. A CNN-Transformer Cooperation Network (CTCNet) is investigated in \cite{10087319} for face hallucination by incorporating a local-global feature cooperation module and a feature refinement module to enhance the local facial details and global facial structure restoration.


From the above, it appears that all existing works do not address the face super-resolution with incremental learning, where the network is trained on one type of faces and extended to other type of faces. This paper proposes the hallucination for face to cartoon incremental learning scenario.

\begin{figure*}[!t]
    \centering
    \includegraphics[width=\textwidth, height=12.0cm]{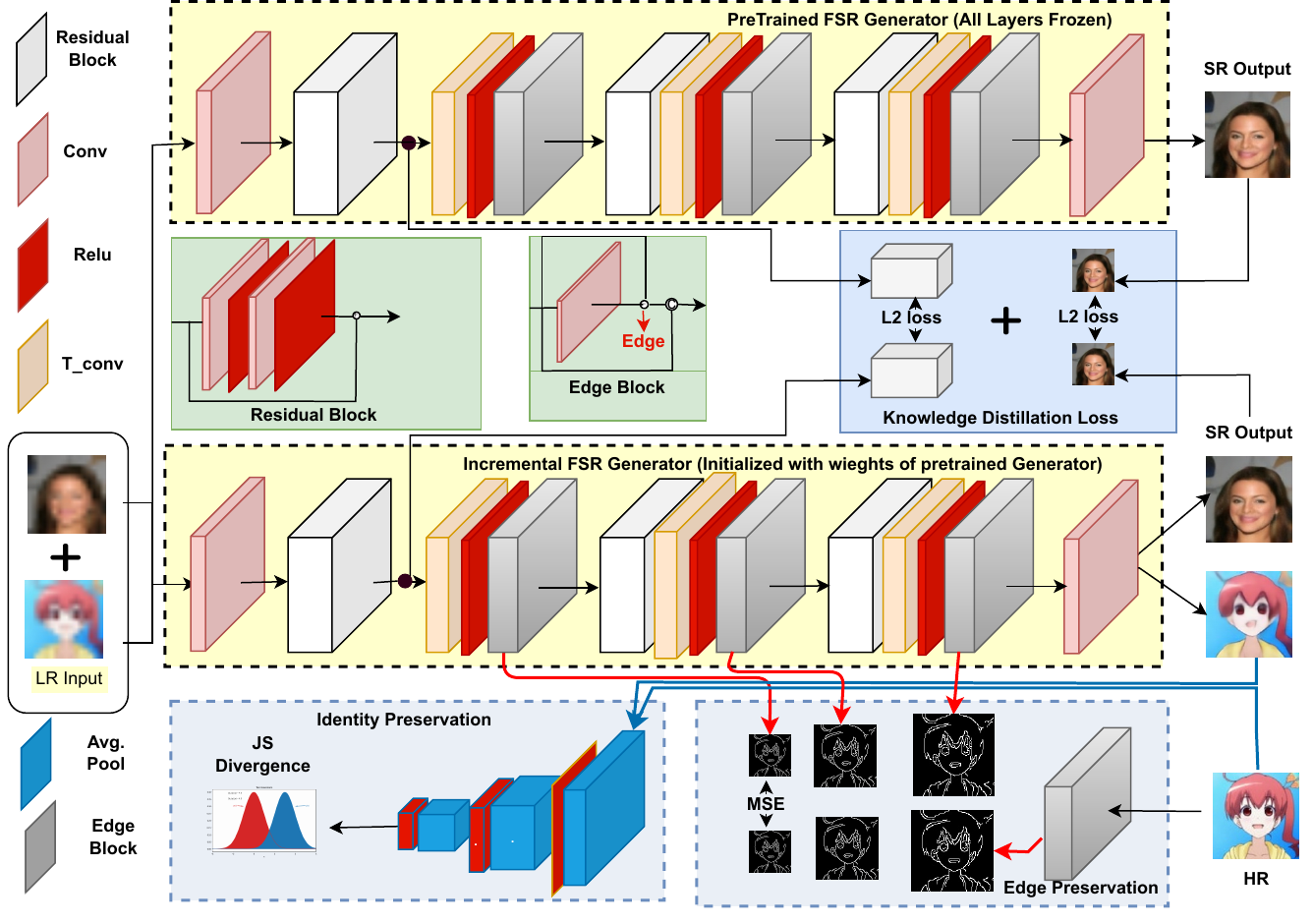}
    \caption{Proposed face to cartoon incremental super-resolution method using knowledge distillation. Conv, ReLU, and T\_conv represent Convolution Layer, ReLU Layer, and Transpose Convolution Layer, respectively. Pre-trained FSR Generator is trained on CelebA Dataset. The incremental FSR Generator is initialized with weights of Pre-trained FSR Generator and trained on combined CelebA and iCartoonFace images using the proposed method.}
    \label{fig:EIP}
\end{figure*}

\section{Proposed Methodology}

\subsection{Problem Description}
We tackle the problem of incremental super-resolution, where we consider a trained model to perform super-resolution in source domain, i.e., $I_{LR}^S \rightarrow I_{SR}^S$, and incrementally train it to perform super-resolution in target domain, i.e. $I_{LR}^T \rightarrow I_{SR}^T$, while retaining the performance for super-resolution in source domain. In the experiments, the source and target domains are considered as human faces from CelebA dataset and cartoon faces from iCartoonFace dataset, respectively.

Let $I_{HR} \in \mathbb{R}^{h_{HR} \times w_{HR} \times c}$ be the high resolution image data, and $I_{LR} \in \mathbb{R}^{h_{LR} \times w_{LR} \times c}$ be the corresponding low resolution image data, where $h_{HR}$ and $w_{HR}$ are height and width of high resolution images, $h_{LR}$ and $w_{LR}$ are height and width of low resolution images, and $c$ is the number of channels. The image degradation function $\phi$ can be described as,
\begin{equation*}
I_{LR} = \phi(I_{HR},\theta),
\end{equation*}
where $\theta$ represents the choice of kernel for down-sampling the image, and the random noise is added to the image to account for random variations during the down-sampling.

Face super-resolution (FSR) is the inverse process of image degradation for down-sampling a high resolution image. Mathematically, it is expressed as,
\begin{equation*}
I_{SR} = \phi^{-1}(I_{LR},\delta) = G(I_{LR},\delta),
\end{equation*}
where $G$ is the FSR generator network with parameters $\delta$, and the super-resolved image is represented by $I_{SR}$.

To update the FSR generator using incremental learning, we use a pre-trained FSR generator $G_S$ trained on the source domain images ($I_{LR}^S$, $I_{HR}^S$) as a starting point. Then, we train our Incremental FSR generator $G_T$ on a new dataset which is created using a combination of source domain and target domain images. Using images from a source domain while incrementally training for a target domain allows us to use knowledge distillation. This helps us in combatting catastrophic forgetting problem. Knowledge distillation is achieved by feeding the low resolution images from source domain $I_{LR}^S$ as input to the pre-trained FSR generator $G_S$ as well as the incremental FSR generator $G_T$, simultaneously and comparing their outputs, i.e., $I_{SR,S}^S = G_{S}(I_{LR}^S)$ and $I_{SR,T}^S = G_{T}(I_{LR}^S)$, using L2 loss. It Incentivizes the incremental FSR generator $G_T$ to maintain its performance on the source domain task. The proposed method is illustrated in Fig. \ref{fig:EIP}.

The incremental learning process combined with knowledge distillation can be formalized as follows,
\begin{equation}
\delta_{optimal} = \underset{\delta_T}{\operatorname{argmin}} \left[ 
\begin{aligned}
&\mathcal{L}_T(I_{HR}^T, G_T(I_{LR}^T, \delta_T)) +\\
& \lambda \mathcal{L}_{kd}(G_T(I_{LR}^S, \delta_T), G_S(I_{LR}^S, \delta_S))
\end{aligned}
 \right],
\end{equation}
where $\mathcal{L}_T$ is the loss function measuring the difference between the super-resolved images $I_{SR,T}^T = G_T(I_{LR}^T, \delta_T)$ and the high-resolution images $I_{HR}^T$ in the target domain.
$\mathcal{L}_{kd}$ is the knowledge distillation loss function, which measures the difference between the outputs of the incremental FSR generator $G_T$ and the pre-trained FSR generator $G_S$ when fed with low-resolution images $I_{LR}^S$ from the source domain.
$\delta_S$ represents the parameters of the pre-trained FSR generator $G_S$ which are obtained by training from scratch on source domain. 
$\delta_T$ represents the parameters of the incremental FSR generator $G_T$ that are first initialized as $\delta_S$ and then updated to minimize the combined loss.

The goal of the above objective is to update the parameters $\delta_T$ of the incremental FSR generator $G_T$ in a way that minimizes the loss functions for target domain images, while also ensuring that the knowledge distillation loss between $G_T$ and $G_S$ is minimized. It should be noted that $\delta_S$ is kept frozen while we train the incremental FSR generator $G_T$ in an incremental fashion.
The regularization coefficient $\lambda$ balances the importance of knowledge distillation loss in the overall objective.

\subsection{Knowledge Distillation}
The proposed method utilizes a pre-trained Facial Super-Resolution (FSR) generator $G_S$, as depicted in Fig. \ref{fig:EIP}, which has been trained on source domain images from the CelebA dataset. To ensure the stability of $G_S$ during the training process, all layers of $G_S$ are frozen. This pre-trained FSR generator is then employed for knowledge distillation to mitigate the issue of catastrophic forgetting.
Another generator, denoted as $G_T$ (Incremental FSR generator), as shown in Fig. \ref{fig:EIP}, is initialised using weights of $G_S$ and incrementally trained on target domain images from the iCartoonFace dataset and a small subset of images from the source domain CelebA dataset. Low resolution images from source domain $I_{LR}^S$ are given as input to $G_S$ and $G_T$, the generated super-resolved images $I_{SR,S}^S$ and $I_{SR,T}^S$ are then used for knowledge distillation. Specifically, knowledge distillation is performed by computing the L2 loss between $I_{SR,S}^S$ and $I_{SR,T}^S$. Basically, this loss encourages the generator $G_T$ to match the output of generator $G_S$ for the source domain images, thereby preventing the forgetting of source domain knowledge by the network $G_T$.

Moreover, the outputs from bottleneck layers of $G_S$ and $G_T$ for low-resolution images from source domain $I_{LR}^S$ are also compared using the L2 loss. This comparison ensures that the bottleneck representation of $G_T$, which is actively being trained on inputs from the target domain, does not deviate significantly from the bottleneck representation of $G_S$, which was trained on inputs from the source domain.
The knowledge distillation loss is expressed as follows,
\begin{equation}
\mathcal{L}_{kd} = \mathcal{L}_{R}(I_{SR,S}^S,I_{SR,T}^S) + \mathcal{L}_{F}(h_t,h_s),
\label{eq:loss_kd}
\end{equation}
where $I_{SR,S}^S$ and $I_{SR,T}^S$ represent the outputs of the $G_S$ and $G_T$, respectively, while $h_s$ and $h_t$ indicate the hidden layer responses from the $G_T$ and $G_S$, respectively for the source domain images. $\mathcal{L}_{R}$ represents the response-based loss, computed on the output logits, and $\mathcal{L}_{F}$ denotes the feature loss, computed from the hidden layer.

\subsection{Edge Block}
Traditional generative networks can lead to blurry images as high-frequency edge information is often lost during image resizing and re-scaling.  Edge information can be used to improve the quality of images generated by face-generative networks by providing additional guidance to the generator during training. The edge block allows the generator to better preserve the fine details of the input images, by capturing the structural differences between two images rather than individual pixels. In addition, incorporating edge information can help to combat the disadvantages of using L2 loss, which is sensitive to changes in individual pixels and causes the blurry generated images.

\begin{figure}[!t]
  \centering
  \includegraphics[width=0.95\columnwidth]{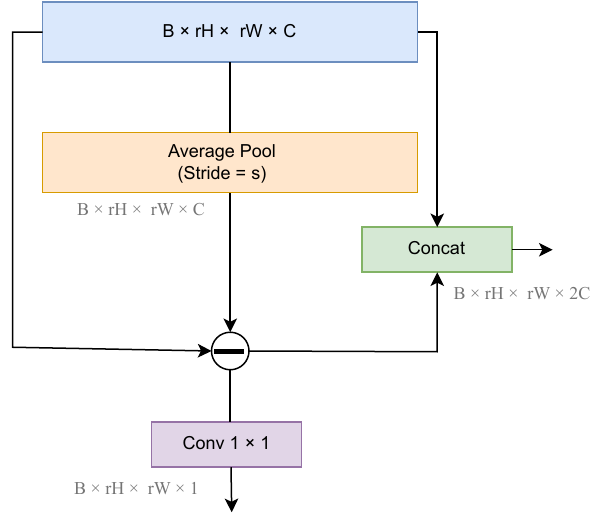}
    \caption{Edge block having an edge extraction layer. Here, $B$ is the batch size, $H$ is the height of tensor, $W$ is the width of tensor, $c$ is the number of channels in tensor, $r$ is the scaling factor and $s$ is the stride variable.}
  \label{fig:Edge}
\end{figure}

The edge block, as depicted in Fig. \ref{fig:Edge}, plays a key role in leveraging edge information within the facial resolution network. This block is designed to be computationally efficient and consists of a low-pass filter constructed using an average pooling layer. The edge block uses a variable kernel size with stride 1 and ``SAME" padding to retain the spatial resolution. The kernel size used during the average pool step depends on the size of input image, i.e., kernel size as ($5, 7, 10$) is used for input tensor of size ($32\times32$, $64\times64$, $128\times128$), respectively. By applying the low-pass filter, a blurred version of the original image is obtained. The difference between the blurred image and the original image yields the edge map, which highlights the edges present in the image. The edge map is then concatenated with the original image and propagated to the subsequent layer. To reduce the number of edge maps to a single-channel representation, a pointwise convolution operation is employed. This single-channel edge map is subsequently compared with the edge map of the high-resolution image, which is obtained using the Canny edge detector \cite{4767851} by applying adaptive threshold selection \cite{6885761}.

Let \(E_{HR}\) be the edge map of the high-resolution image and \(E_{SR}\) be the edge map of the generated super-resolved image.
The edge loss (\(\mathcal{L}_{edge}\)) is computed by comparing the generated edge map (\(E_{SR}\)) with the ground truth edge map (\(E_{HR}\)). Therefore, the edge loss (\(\mathcal{L}_{edge}\)) can be defined as:
\begin{equation}    
\mathcal{L}_{edge} = \frac{1}{H \times W} \sum_{i=1}^{H} \sum_{j=1}^{W} (E_{SR}(i,j) - E_{HR}(i,j))^2,
\label{loss_edge}
\end{equation}
where \(H\) and \(W\) represent the height and width of the edge maps, respectively.
Minimizing the edge loss encourages the generated image to have similar edge structures as the high-resolution ground truth image, leading to enhanced sharpness and preservation of fine details.

\subsection{Generator Architecture}
The generator network consists of three similar modules with a convolution layer at the beginning and at the end. Each module is made up of a residual block, transpose convolution layer, ReLU activation function and an edge block. The edge block is shown in Fig. \ref{fig:Edge}. The residual block is made up of two convolution layers, each followed by a ReLU activation function. The transpose convolution performs a $2\times$ up-sampling. Overall, the network performs $8\times$ up-sampling. In Fig. \ref{fig:EIP}, refer to Incremental FSR Generator for detailed visualization. The generator is trained with adversarial loss and other losses described in this paper.

\begin{figure}[!t]
  \centering
  \includegraphics[width=\columnwidth]{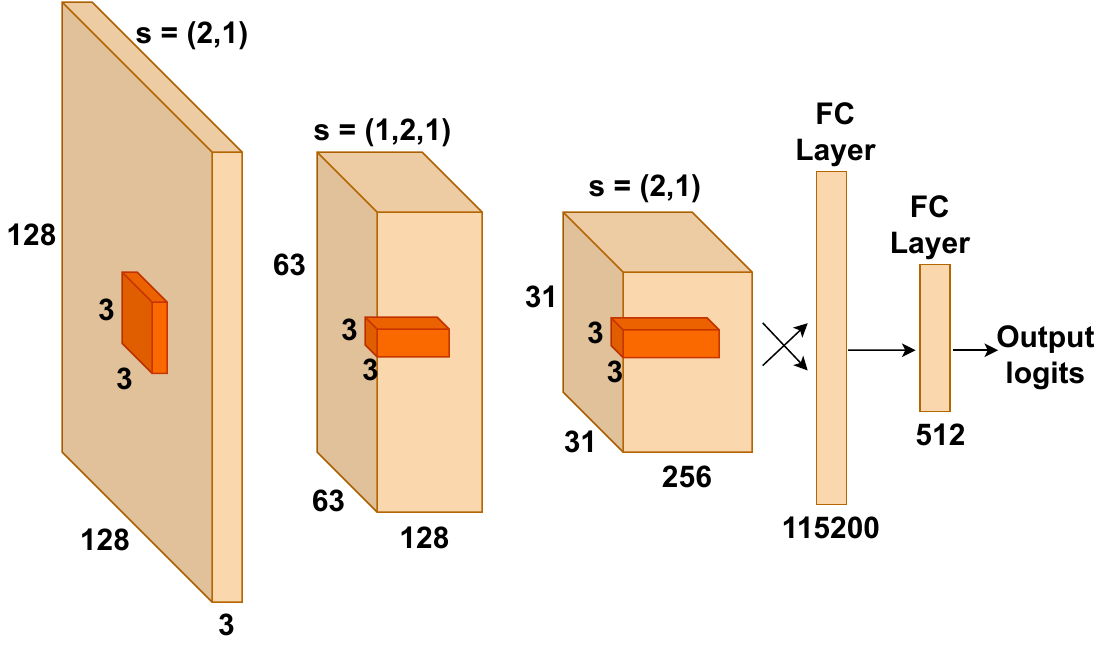}
      \caption{A schematic diagram of Discriminator architecture. Here, $s$ indicates the stride and the list of numbers adjacent to $s$ indicates the stride of convolution layers grouped with the same number of output channels.}
  \label{fig:DISC}
\end{figure}

\subsection{Discriminator Architecture}
The discriminator is a CNN used to distinguish between real and generated images. It is illustrated in Fig. \ref{fig:DISC}. The network consists of seven convolution layers with leaky ReLU activation in the first six layers and two fully-connected layers. The input to the network is a super-resolution image which is passed through the network to determine if it is a real or a generated image. Initially, the input image passes through the first convolution layer, which has a kernel size of $3 \times 3$, a stride of $1$, and $128$ output channels. The output of this layer is then passed through another convolution layer, which has a kernel size of $3 \times 3$, a stride of $2$, and $128$ output channels. The same process is repeated for two more convolution layers, each with $256$ output channels. After these layers, the image is passed through two more convolution layers, with $256$ and $512$ output channels, respectively. Finally, the image is passed through one last convolution layer, with $512$ output channels. The output of this layer is then flattened and passed through two fully-connected layers to determine whether the input image is a real or a generated high-resolution image. The leaky ReLU activation function is used for all convolution layers with a slope of $0.2$ for negative activations. The generator and discriminator networks are trained with an adversarial loss function to classify real and generated images.

\subsection{Objective Function}
The objective function for Face to Cartoon Incremental Super-Resolution using Knowledge Distillation task is given by loss $\mathcal{L}_{ISR-KD}$ as,
\begin{equation}
\begin{split}
\mathcal{L}_{ISR-KD} = \lambda_{kd}\mathcal{L}_{kd} + \lambda_{edge}\mathcal{L}_{edge} + \lambda_{ad}\mathcal{L}_{ad} \\
+ \lambda_{lce}\mathcal{L}_{lce} 
 + \lambda_{id}\mathcal{L}_{id} + \lambda_{rl}\mathcal{L}_{rl}, 
\end{split}
\end{equation}
where $\mathcal{L}_{kd}$, $\mathcal{L}_{edge}$, $\mathcal{L}_{ad}$, $\mathcal{L}_{lce}$, $\mathcal{L}_{id}$ and $\mathcal{L}_{rl}$ are knowledge distillation loss, edge loss, adversarial loss, luminance-chrominance error, identity loss and reconstruction loss, respectively. $\lambda_{kd}$ represents the combined effect of ${L}_{R}$ and ${L}_{F}$ on $\mathcal{L}_{kd}$ as shown in Eq. (\ref{eq:loss_kd}). $\lambda$'s are the hyper parameters used as the weights for different loss functions. $\mathcal{L}_{edge}$ loss is defined in Eq. (\ref{loss_edge}), respectively. $\mathcal{L}_{ad}$, $\mathcal{L}_{lce}$, $\mathcal{L}_{id}$ and $\mathcal{L}_{rl}$ losses are described in the rest of this subsection.

\subsubsection{Adversarial Loss}
The adversarial loss function is defined as,
\begin{equation}
\begin{split}
L_{ad} =  \mathbb{E}_{x \sim p(I_{HR}^T)} [\log D_T(x)] + \\ \mathbb{E}_{z \sim p(I_{LR}^T)} [\log (1 - D_T(G_T(z)))].
\label{GAN_loss}
\end{split}
\end{equation}
The aim of the Generator is to minimize the second term of equation $L_{ad}$ as it wants to fool the discriminator by predicting the generated samples as real. The aim of the Discriminator is to maximize $L_{ad}$ to make sure that the discriminator can accurately distinguish between samples coming from the probability distribution of high resolution images $p(I_{HR}^T)$ and the probability distribution of hallucination images $p(I_{SR,T}^T)$.

\subsubsection{Luminance-Chrominance Error}
Luminance-chrominance error occurs in image super-resolution tasks due to the mismatch between luminance (brightness) and chrominance (color) components. 

The difference between two images in the Luminance-chrominance space (i.e., YCbCr) is calculated as,
\begin{equation}
\mathcal{L}_{lce} = \sqrt{\Delta Y^2 + \Delta Cb^2 + \Delta Cr^2},
\end{equation}
where $\Delta Y$, $\Delta Cb$ and $\Delta Cr$ are the difference between super-resolution and high-resolution images in $Y$, $Cb$ and $Cr$ channels, respectively. Minimizing $\mathcal{L}_{lce}$ improves the Luminance-chrominance preservation in the super-resolution images.


\subsubsection{Identity Loss}
We use the identity loss as the JS divergence which measures the similarity between two probability distributions. The generated and ground truth images are passed through a pre-trained Inception-V1 network to extract 512-class encoded vectors, denoted as $V_{SR}$ and $V_{HR}$, respectively. The identity loss is calculated as,
\begin{equation}
\mathcal{L}_{id} = \frac{1}{2} KL(V_{SR} | M) + \frac{1}{2} KL(V_{HR} | M),
\end{equation}
where $KL$ is the Kullback-Leibler divergence and $M$ is the average distribution of $V_{SR}$ and $V_{HR}$ given as, $M = (V_{SR} + V_{HR})/2$. The identity loss as JS divergence assesses the similarity between high-level features of generated and ground truth images, measuring how closely the generated output aligns with the high-resolution images.

\subsubsection{Reconstruction Loss}
The reconstruction loss is computed as mean squared error (MSE) between the pixel values of the generated image and the corresponding pixel values of the ground truth image. Mathematically, it can be expressed as,
\begin{equation}
    {L_{rl}} = \frac{1}{N} \sum_{i=1}^{N} (I_{SR_i} - I_{HR_i})^2,
\end{equation}
where $N$ represents the total number of pixels in the images, and $I_{SR_i}$ and $I_{HR_i}$ denote the pixel values for the $i^{th}$ pixel in the generated and ground truth images, respectively.



\begin{table*}[!t]
\caption{The Experimental results of the proposed ISR-KD model on different dataset settings. The \% change is obtained by comparing with performance of the model on CelebA dataset trained from scratch.}
\label{tab:combined-results}
\centering
\resizebox{\textwidth}{!}{
\begin{tabular}{|p{0.15\textwidth}|p{0.28\textwidth}|p{0.09\textwidth}|p{0.13\textwidth}|p{0.12\textwidth}|p{0.13\textwidth}|}
\hline
\textbf{Dataset} & \textbf{ISR-KD Setting} & \textbf{Test Dataset} & \textbf{PSNR} & \textbf{SSIM} & \textbf{FID} \\ \hline
\multirow{2}{*}{CelebA (From Scratch)} & \multirow{2}{*}{NA} & CelebA & 24.2420 & 0.7097 & 44.3340 \\ \cline{3-6}
& & iCartoonFace & 20.5817 & 0.5781 & 113.3130 \\ \hline
\multirow{2}{*}{Cartoon-CelebA-1} & \multirow{2}{*}{0 CelebA and 20,000 Cartoon images} & CelebA & 23.5397 ($\downarrow$\SI{2.90}{\percent}) & 0.6817 ($\downarrow$\SI{3.95}{\percent}) & 74.3212 ($\downarrow$\SI{67.64}{\percent}) \\ \cline{3-6}
& & iCartoonFace & 20.6933 ($\uparrow$\SI{0.54}{\percent}) & 0.5991 ($\uparrow$\SI{3.63}{\percent}) & 98.8749 ($\uparrow$\SI{12.74}{\percent}) \\ \hline
\multirow{2}{*}{Cartoon-CelebA-2} & \multirow{2}{*}{10,000 CelebA and 20,000 Cartoon images} & CelebA & 23.9387 ($\downarrow$\SI{1.25}{\percent}) & 0.7007 ($\downarrow$\SI{1.27}{\percent}) & 59.3153 ($\downarrow$\SI{33.79}{\percent}) \\ \cline{3-6}
& & iCartoonFace & 20.7170 ($\uparrow$\SI{0.66}{\percent}) & 0.6007 ($\uparrow$\SI{3.91}{\percent}) & 98.0944 ($\uparrow$\SI{13.43}{\percent}) \\ \hline
\multirow{2}{*}{Cartoon-CelebA-3} & \multirow{2}{*}{20,000 CelebA and 20,000 Cartoon images} & CelebA & 23.9526 ($\downarrow$\SI{1.19}{\percent}) & 0.7005 ($\downarrow$\SI{1.30}{\percent}) & 61.3342 ($\downarrow$\SI{38.35}{\percent}) \\ \cline{3-6}
& & iCartoonFace & 20.7170 ($\uparrow$\SI{0.66}{\percent}) & 0.5998 ($\uparrow$\SI{3.75}{\percent}) & 99.3709 ($\uparrow$\SI{12.30}{\percent}) \\ \hline
\multirow{2}{*}{Cartoon-CelebA-4} & \multirow{2}{*}{10,000 CelebA and 50,000 Cartoon images} & CelebA & 24.0124 ($\downarrow$\SI{0.95}{\percent}) & \begin{tabular}[c]{@{}l@{}}0.7047 ($\downarrow$\SI{0.70}{\percent})\end{tabular} & \begin{tabular}[c]{@{}l@{}}57.2367 ($\downarrow$\SI{29.10}{\percent})\end{tabular} \\ \cline{3-6}
& & iCartoonFace & 20.8310 ($\uparrow$\SI{1.21}{\percent}) & 0.6092 ($\uparrow$\SI{5.38}{\percent}) & 87.6397 ($\uparrow$\SI{22.66}{\percent}) \\ \hline

\multirow{2}{*}{Cartoon-CelebA-5} & \multirow{2}{*}{20,000 CelebA and 50,000 Cartoon images} & CelebA & 24.1403 ($\downarrow$\SI{0.42}{\percent}) & \begin{tabular}[c]{@{}l@{}}0.7096 ($\downarrow$\SI{0.01}{\percent})\end{tabular} & \begin{tabular}[c]{@{}l@{}}58.1901 ($\downarrow$\SI{31.25}{\percent})\end{tabular} \\ \cline{3-6}
& & iCartoonFace & 20.8387 ($\uparrow$\SI{1.25}{\percent}) & 0.6072 ($\uparrow$\SI{5.03}{\percent}) & 91.8344 ($\uparrow$\SI{18.96}{\percent}) \\ \hline
\end{tabular}}
\end{table*}

\section{Experimental Settings}

\subsection{Datasets}
For experimental analysis, CelebA \cite{liu2015faceattributes} and iCartoonFace \cite{zheng2020cartoon} datasets are used in this paper. The CelebA dataset contains 202,599 face images from 10,177 identities. Whereas, the iCartoonFace dataset contains 389,678 cartoon face images from 5,013 identities. The CelebA dataset is used as the source domain and iCartoonFace dataset is used as the target domain. In this paper, we experiment with five different combinations by considering varying number of images randomly from CelebA and iCartoonFace datasets as detailed in Table \ref{tab:combined-results}. The five dataset combinations are Cartoon-CelebA-1, Cartoon-CelebA-2, Cartoon-CelebA-3, Cartoon-CelebA-4 and Cartoon-CelebA-5.

\subsection{Experimental Setup}
The data augmentation is performed by center cropping to a size of $178 \times 178$ pixels, resizing to a size of $128 \times 128$ pixels, horizontal flipping with a probability of $0.5$, and rotating with $90$ and $270$ degrees. The generator and discriminator are trained using the Adam optimizer with learning rate $1e-4$ and Epsilon $1e-8$. $Beta1$ and $Beta2$ for Generator are $0.9$ and $0.999$, respectively. However, for Discriminator they are $0.5$ and $0.9$, respectively. 
The model is trained for $100$ epochs. The model is trained and tested on an Nvidia Quadro RTX 6000 GPU using the TensorFlow framework. 
The values of $\mathcal{L}_{R}$, $\mathcal{L}_{F}$, and $\lambda_{edge}$ hyperparameters are $5$, $0.01$, and $0.3$. However, $\lambda_{ad}$, $\lambda_{lce}$, $\lambda_{id}$, and $\lambda_{rl}$ are set to $1$.

\begin{figure*}[!t]
    \centering
    \includegraphics[width=\textwidth, height=10cm]{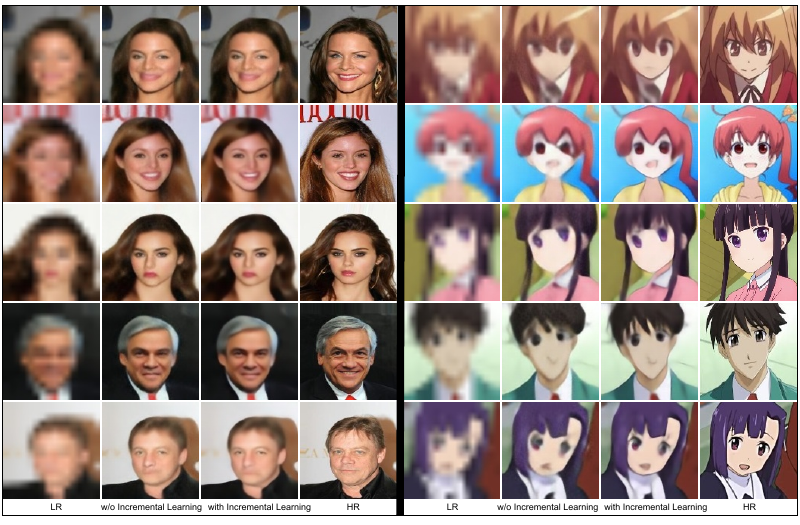}
    \caption{The generated samples depicting the visual effects of using incremental learning in combination with knowledge distillation for facial super-resolution task. The left half of the image contains the results for CelebA dataset (Source Domain). The right half shows the results after incrementally training on the iCartoonFace dataset (Target Domain).}
    \label{fig:visual_effects}
\end{figure*}

\begin{figure*}[!t]
    \centering
    \includegraphics[width=\textwidth, height=12.5cm]{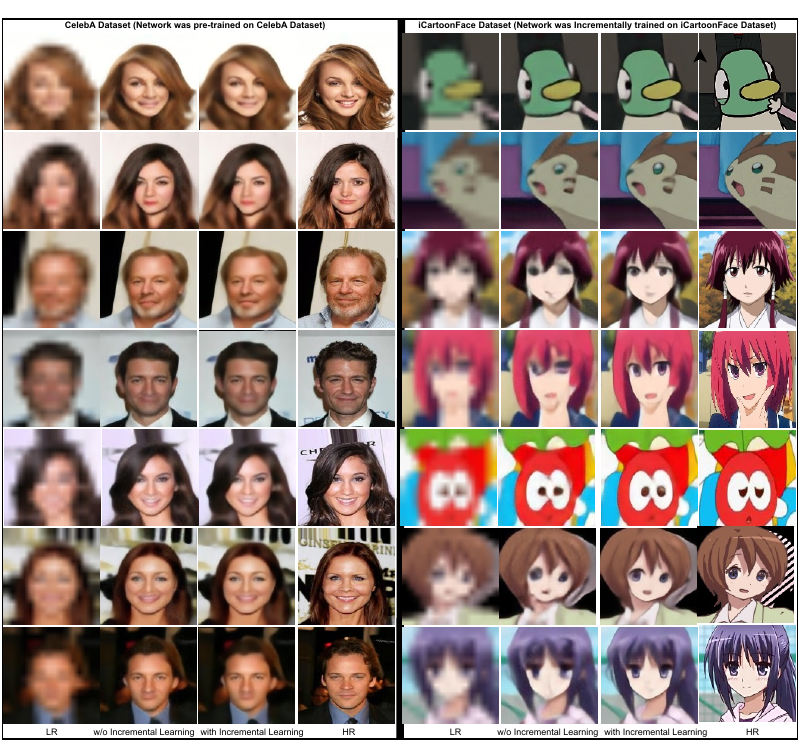}
    \caption{The additional visual results. The generated samples depicting the visual effects of using incremental learning in combination with knowledge distillation for facial super-resolution task. The left half of the image contains the results for CelebA dataset (Source Domain). The right half shows the results after incrementally training on the iCartoonFace dataset (Target Domain).}
    \label{fig:additional}
\end{figure*}

\begin{table}[!t]
    \caption{Similarity between the samples generated without and with incremental learning using the proposed approach.}
    \centering
    \begin{tabular}{|c|c|c|c|c|}
    \hline
        \textbf{Test Results Images} & \textbf{PSNR} & \textbf{SSIM} & \textbf{FID} \\ \hline
        CelebA Test Set & 33.2640 & 0.9425 & 18.7007 \\ \hline
        iCartoon Test Set & 28.8210 & 0.8850 & 25.3614 \\ \hline
    \end{tabular}
    \label{tab:qualitative_quantitative}
\end{table}

\section{Experimental Results and Discussion}

\subsection{Quantitative Results}
Table \ref{tab:combined-results} presents the performance of the proposed ISR-KD network. In case of Cartoon-CelebA-1 dataset combination, where knowledge distillation is not employed as this dataset does not contain any CelebA face images, we observe an increase in the performance on the iCartoonFace test set, but a significant drop in the performance on the CelebA dataset. This decline can be attributed to the catastrophic forgetting problem, which occurs when a model forgets previously learned knowledge while training on new data.
To mitigate this problem, we introduced incremental hallucination using knowledge distillation by re-feeding some images from the source domain (i.e., CelebA face images) to a pre-trained generator as shown for Cartoon-CelebA-2, Cartoon-CelebA-3, Cartoon-CelebA-4, and Cartoon-CelebA-5 datasets. This additional step introduces a knowledge distillation loss, which helps to overcome catastrophic forgetting. As a result, the performance on the CelebA test set is almost retained, while simultaneously improving the performance on the iCartoonFace test set. An improvement of 5.03\% on iCartoonFace dataset and a drop of only 0.01\% on CelebA dataset is observed in Table \ref{tab:combined-results} in terms of SSIM on Cartoon-CelebA-5 training dataset setting, which has 20,000 images from CelebA and 50,000 images from iCartoonFace. We note that the number of samples from Cartoon images also plays an important role as the results using Cartoon-CelebA-5 setting are better than the Cartoon-CelebA-3 setting, in spite of having the same number of CelebA samples. Hence, the proposed approach can effectively deal with the catastrophic forgetting issue in the context of super-resolution.

\subsection{Qualitative Results}
Figure \ref{fig:visual_effects} and \ref{fig:additional} illustrate the visual results achieved by combining Incremental training and knowledge distillation. The ISR-KD model was first trained on the entire CelebA training set before undergoing incremental training on a small subset of the iCartoonFace dataset. The left portion of the image displays results for the CelebA dataset. Given that the model was initially trained on the CelebA dataset and subsequently incrementally trained on the iCartoonFace dataset, one might anticipate a substantial decline in the quality of super-resolved (SR) images after incremental training on the iCartoonFace dataset, an occurrence referred to as catastrophic forgetting. However, by employing knowledge distillation, we were able to counteract these effects, and the results clearly indicate that the images with and without incremental learning are nearly indistinguishable. The quality of images super-resolved from the iCartoonFace dataset improved after incremental training. In order to show the degree of degradation on source images and degree of improvement on target images, we compute the average similarity between the generated samples without and with incremental learning on both CelebA and iCartoon test sets in terms of PSNR, SSIM and FID in Table \ref{tab:qualitative_quantitative}. The higher PSNR \& SSIM and smaller FID on CelebA test set shows higher similarity between the generated images using without and with incremental learning, which means low degradation. The vice-versa is observed on iCartoon test set which shows more dissimilarity means high improvement in the generated samples after applying the incremental learning.

\subsection{Ablation Study on Loss Hyperparameters}
Table \ref{tab:Hyperparameter-fine-tune} presents the results of different loss hyperparameter settings on the Cartoon-CelebA-2 dataset. We focus on adjusting $\mathcal{L}_{R}$ and $\mathcal{L}_{F}$ values in Eq. (\ref{eq:loss_kd}), while keeping $\lambda{edge}$ fixed at $0.3$ and other hyperparameters set to $1$. 
The $\mathcal{L}_{R}$ = 5 and $\mathcal{L}_{F}$ = 0.01 are used in other experiments as this setting yields the best performance on the iCartoonFace dataset. This choice of hyperparameters shows a good trade-off between maintaining high performance on the CelebA dataset while enhancing the performance on the iCartoonFace dataset. 

\begin{table}
\caption{Performance comparison of various $\mathcal{L}_{R}$ and $\mathcal{L}_{F}$ loss hyperparameter settings in Eq. (\ref{eq:loss_kd}) on the Cartoon-CelebA-2 dataset with $\lambda_{edge}$ fixed at 0.3 and all other hyperparameters set to 1.}
\label{tab:Hyperparameter-fine-tune}
\resizebox{\columnwidth}{!}{
\begin{tabular}{|l|l|l|l|l|}
\hline
\textbf{Hyper-parameters} & \textbf{Test Dataset} & \textbf{PSNR} & \textbf{SSIM} & \textbf{FID} \\ \hline
\multirow{2}{*}{\begin{tabular}[c]{@{}l@{}}$\mathcal{L}_{R}$ = 15 \\ $\mathcal{L}_{F}$ = 0.04\end{tabular}} & CelebA & 24.0721 & 0.7086 & 58.0514 \\ \cline{2-5}
& iCartoonFace & 20.6202 & 0.5920 & 108.8132 \\ \hline
\multirow{2}{*}{\begin{tabular}[c]{@{}l@{}}$\mathcal{L}_{R}$ = 10 \\ $\mathcal{L}_{F}$ = 0.08\end{tabular}} & CelebA & 24.1119 & 0.7082 & 57.1407 \\ \cline{2-5}
& iCartoonFace & 20.7152 & 0.5962 & 103.8281 \\ \hline
\multirow{2}{*}{\begin{tabular}[c]{@{}l@{}}$\mathcal{L}_{R}$ = 5 \\ $\mathcal{L}_{F}$ = 0.08\end{tabular}} & CelebA & 24.0707 & 0.7070 & 57.7262 \\ \cline{2-5}
& iCartoonFace & 20.7109 & 0.5982 & 100.7698 \\ \hline
\multirow{2}{*}{\begin{tabular}[c]{@{}l@{}}$\mathcal{L}_{R}$ = 5 \\ $\mathcal{L}_{F}$ = 0.04\end{tabular}} & CelebA & 24.0941 & 0.7069 & 58.0286 \\ \cline{2-5}
& iCartoonFace & 20.7397 & 0.5991 & 100.931 \\ \hline
\multirow{2}{*}{\begin{tabular}[c]{@{}l@{}}$\mathcal{L}_{R}$ = 5 \\ $\mathcal{L}_{F}$ = 0.01\end{tabular}} & CelebA & 23.9387 & 0.7007 & 59.3153 \\ \cline{2-5}
& iCartoonFace & 20.717 & 0.6007 & 98.0944 \\ \hline
\end{tabular}}
\end{table}

\begin{table}[!t]
    \caption{Cross-dataset analysis by first training on Cartoon images from scratch and then incrementally learn on CelebA images using the proposed method.}
    \centering
    \resizebox{\columnwidth}{!}{
    \begin{tabular}{|p{0.254\columnwidth}|p{0.219\columnwidth}|p{0.12\columnwidth}|p{0.10\columnwidth}|p{0.12\columnwidth}|}
    \hline
        \textbf{Train Dataset} & \textbf{Test Dataset} & \textbf{PSNR} & \textbf{SSIM} & \textbf{FID} \\ \hline
        Cartoon & iCartoonFace & 20.8648 & 0.6055 & 94.0667 \\ \cline{2-5}
        (From Scratch) & CelebA & 23.6665 & 0.6834 & 75.2700 \\ \hline
        Incremental & iCartoonFace & 20.8364 & 0.6061 & 90.7447 \\ \cline{2-5}
        Learning & CelebA & 24.1532 & 0.7074 & 56.3191 \\ \hline
    \end{tabular}}
    \label{tab:cross_dataset}
\end{table}

\subsection{Cross-Dataset Analysis}
We also perform the cross-dataset analysis by first training the model from scratch on $50,000$ Cartoon images and then applying incremental learning using the proposed approach on a dataset consisting of $20,000$ Cartoon images from source domain and $20,000$ CelebA face images from the target domain. The results reported in Table \ref{tab:cross_dataset} confirm that the proposed ISR-KD improves the performance on CelebA while maintaining the similar performance on iCartoon dataset. This analysis also points out that performing super-resolution on cartoon face images is a difficult problem as compared to normal face images.



\begin{table}[!t]
\caption{Results for the Extended ISR-KD network when trained on Cartoon-CelebA-2 dataset setting.}
\label{tab:extended}
\centering
\resizebox{\columnwidth}{!}{
\begin{tabular}{|l|l|l|l|}
\hline
\textbf{Test Dataset} & \textbf{PSNR}        & \textbf{SSIM}        & \textbf{FID}          \\ \hline
CelebA                & 24.19 ($0.23\downarrow$) & 0.71 ($0.31\uparrow$) & 58.42 ($31.76\uparrow$) \\ \hline
iCartoonFace              & 20.74 ($0.77\uparrow$) & 0.60 ($3.94\uparrow$) & 98.02 ($13.49\uparrow$) \\ \hline
\end{tabular}}
\end{table}

\subsection{Performance on Extended Network}
In this experiment, the incremental FSR generator network is extended by adding six convolution layers with padding set as SAME to the rear end of the network. The newly added layers are initialized with random weights. The weights of the other layers are initialized from the pre-trained FSR generator (Fig. \ref{fig:EIP}). The extended model is trained on the Cartoon-CelebA-2 dataset. Table \ref{tab:extended} shows that increasing the depth of incremental FSR generator leads to better results for source (CelebA) as well as incremental target (iCartoonFace) facial hallucination tasks.

\section{Conclusion}
In this research paper, we addressed the problem of adapting GANs to new and unseen data in the context of facial hallucination. We combine the incremental learning and knowledge distillation in the proposed ISR-KD framework. The incorporation of knowledge distillation allows the model to retain the performance on previous dataset while enhancing its capability on new dataset. We used the pre-trained GAN-based super-resolution network on the CelebA dataset and incrementally trained it on the combined CelebA and iCartoonFace dataset using our proposed framework. We achieved superior performance on target Cartoon dataset while maintaining the performance on the source CelebA dataset, mitigating the issue of catastrophic forgetting.
Future research directions include investigating different knowledge distillation methods, different networks, and evaluating on larger and more diverse datasets.



{\small
\bibliographystyle{ieee}
\bibliography{Reference}
} 
\end{document}